\documentclass[10pt,twocolumn,letterpaper]{article}

\usepackage{iccv}
\usepackage{times}
\usepackage{epsfig}
\usepackage{graphicx}
\usepackage{amsmath}
\usepackage{amssymb}

\usepackage{url}
\usepackage{booktabs}
\usepackage{amssymb}
\usepackage{bbding}
\usepackage{pifont}
\usepackage{wasysym}
\usepackage{utfsym}
\usepackage{fontawesome}
\usepackage{multirow}
\usepackage{colortbl}
\usepackage{xcolor}
\usepackage{array} 
\usepackage{makecell}

\usepackage[pagebackref=true,breaklinks=true,colorlinks,bookmarks=false]{hyperref}
\iccvfinalcopy 

\def\iccvPaperID{****} 

\ificcvfinal\fi

\begin{document}

\title{\vspace{-0.6cm}T2I-Adapter: Learning Adapters to Dig out More Controllable Ability for Text-to-Image Diffusion Models}

\vspace{-1cm}
\author{
	Chong Mou$^{*1,2}$ \hspace{9pt} Xintao Wang$^\dagger$$^{2}$ \hspace{9pt} Liangbin Xie$^{*2,3,4}$ \hspace{9pt} Yanze Wu$^{2}$ \hspace{9pt} Jian Zhang$^\dagger$$^{1}$ \hspace{9pt} \\
    Zhongang Qi$^{2}$ \hspace{9pt} Ying Shan$^{2}$  \hspace{9pt}Xiaohu Qie$^{2}$ \\
	\vspace{-0.05cm}
\small$^1$Peking University Shenzhen Graduate School \hspace{5pt} 
\small$^2$ARC Lab, Tencent PCG \hspace{5pt}\small$^3$University of Macau \hspace{5pt}\small$^4$Shenzhen Institute of Advanced Technology\\
\url{https://github.com/TencentARC/T2I-Adapter}
}
\maketitle 
\let\thefootnote\relax\footnotetext{$^*$ Interns in ARC Lab, Tencent PCG \hspace{3pt} $^\dagger$ Corresponding author.
}

\begin{abstract}
The incredible generative ability of large-scale text-to-image (T2I) models has demonstrated strong power of learning complex structures and meaningful semantics. However, relying solely on text prompts cannot fully take advantage of the knowledge learned by the model, especially when flexible and accurate controlling (\textit{e.g.}, color and structure) is needed. In this paper, we aim to ``dig out" the capabilities that T2I models have implicitly learned, and then explicitly use them to control the generation more granularly. Specifically, we propose to learn simple and lightweight \textbf{T2I-Adapters} to align internal knowledge in T2I models with external control signals, while freezing the original large T2I models. In this way, we can train various adapters according to different conditions, achieving rich control and editing effects in the color and structure of the generation results. Further, the proposed T2I-Adapters have attractive properties of practical value, such as composability and generalization ability. Extensive experiments demonstrate that our T2I-Adapter has promising generation quality and a wide range of applications.

\end{abstract}

\section{Introduction}

Thanks to the training on massive data and huge computing power, text-to-image (T2I) generation~\cite{t2i1,ldm,glid,t2i2,t2i3,t2i4,t2i5}, which aims to generate images conditioned on a given text/prompt, has demonstrated strong generation ability.
The generation results usually have rich textures, clear edges, reasonable structures, and meaningful semantics. 
This phenomenon potentially indicates that T2I models can actually well capture information of different levels in an \textit{implicit} way, from low level (\textit{e.g.}, textures), middle level (\textit{e.g.}, edges) to high level (\textit{e.g.,} semantics).

Although promising synthesis quality can be achieved, it heavily relies on well-designed prompts~\cite{good_prompt1, good_prompt2}, and the generation pipeline also lacks flexible user control capability that can guide the generated images to realize users' ideas accurately.
For an unprofessional user, the generated results are usually uncontrolled and unstable. 
For example, the recently proposed Stable Diffusion (SD)~\cite{ldm} can not perform well in some imaginative scenarios, \textit{e.g.}, \textit{``A car with flying wings"} and \textit{``Iron Man with bunny ears"} as shown in Fig.~\ref{fig:teaser}. 
We believe that this does not mean that T2I models do not have the ability to generate such structures, just that the text cannot provide accurate structure guidance.
%


In this paper, we are curious about whether it is possible  to somehow ``dig out" the capabilities that T2I models have implicitly learned, especially the high-level structure and semantic capabilities, and then explicitly use them to control the generation more accurately. 

%
%

We believe that a small adapter model can achieve this purpose, as it is not learning new generation abilities, but learning a mapping from control information to the internal knowledge in T2I models. In other words, the main problem here is the ``alignment'' issue, \textit{i.e.}, the internal knowledge and external control signal should be aligned.

\begin{figure}[t]
\centering
\small 
\begin{minipage}[t]{\linewidth}
\centering
\includegraphics[width=1\columnwidth]{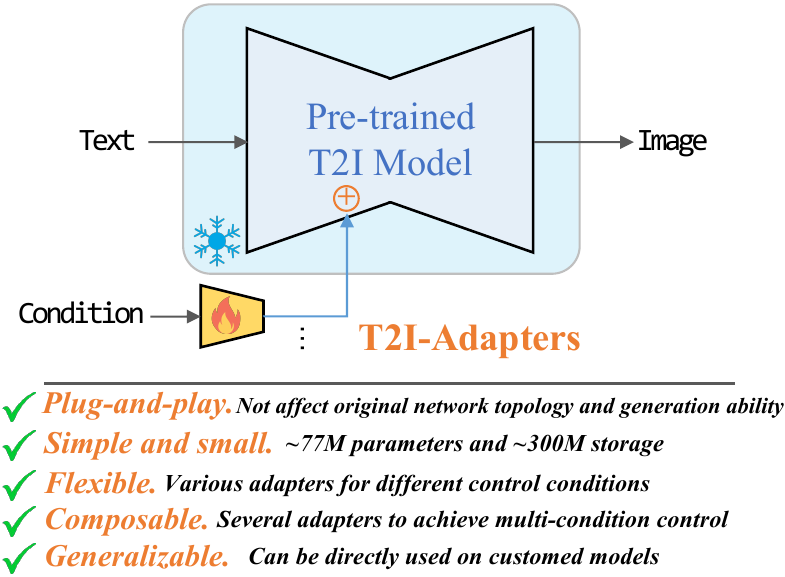}
\end{minipage}
\centering
\caption{Our simple T2I-Adapter can provide extra guidance to pre-trained text-to-image models while not affecting their original generation ability. It also has several attractive properties of practical value. }
\label{mot} 
\end{figure}

Therefore, we propose the T2I-Adapter, which is a lightweight model and can be used to learn this alignment with a relatively small amount of data.
T2I-Adapter provides the pre-trained T2I diffusion models (\textit{i.e.}, SD~\cite{ldm}) with extra guidance.
In this way, we can train various adapters according to different conditions, and they can provide more accurate and controllable generation guidance for the pre-trained T2I models. As shown in Fig.~\ref{mot}, the T2I-Adapters, as extra networks to inject guidance information, have the following properties of practical value:



\begin{itemize}
    \item \textbf{Plug-and-play}. They do not affect the original network topology and generation ability of existing T2I diffusion models (\textit{e.g.}, Stable Diffusion).
    \item \textbf{Simple and small}. They can be easily inserted into existing T2I diffusion models with low training costs, and they only need one inference during the diffusion process. They are lightweight with $\sim\mathbf{77\ M}$ parameters and $\sim\mathbf{300\ M}$ storage space.
    \item \textbf{Flexible}. We can train various adapters for different control conditions, including spatial color control and elaborate structure control.
    \item \textbf{Composable}. More than one adapter can be easily composed to achieve multi-condition control.
    \item \textbf{Generalizable}. Once trained, they can be directly used on custom models as long as they are fine-tuned from the same T2I model. 
\end{itemize}

Our contributions are summarized as follows: 
\textbf{1).} 
We propose T2I-Adapter, a simple, efficient yet effective method to well align the internal knowledge of T2I models and external control signals with a low cost.
\textbf{2).} T2I-Adapter can provide more accurate controllable guidance to existing T2I models while not affecting their original generation ability.
\textbf{3).} Extensive experiments demonstrate that our method works well with various conditions, and these conditions can also be easily composed to achieve multi-condition control. \textbf{4).} The proposed T2I-Adapter also has an attractive generalization ability to work on some custom models and coarse conditions \textit{e.g.}, free-hand style sketch.

\section{Related Work}
\subsection{Image Synthesis and Translation}
The high-dimensional and structural characteristics bring a great challenge to natural image synthesis. Generative adversarial networks (GAN)~\cite{gan} allows efficient sampling in the random distribution and achieve promising synthesis quality. Some other methods (\textit{e.g.}, variational autoencoders~\cite{vae} and flow models~\cite{flow}) are also proposed to construct a more stable optimization process. Most of these early works perform image synthesis in an unconditional way. In contrast to unconditional image synthesis, some conditional strategies are also proposed. The commonly used condition is the image in another domain, \textit{e.g.}, sketch, semantic segmentation map, and keypose. Several conditional GAN methods~\cite{i2i1,i2i2,i2i3,i2i4} are proposed to translate the condition map in other domains to natural images. In addition to the image condition, text~\cite{t2i1,t2i2,t2i3,t2i4} is also an important condition, which aims to generate an image conditioned on a text description. Most of these methods treated different conditions independently with specific training. Some recent attempts~\cite{mul1} also explore performing image synthesis with multi-modal conditions. 

\subsection{Diffusion Models}
In recent years, the diffusion model~\cite{diff} has achieved great success in the community of image synthesis. It aims to generate images from Gaussian noise via an iterative denoising process. Its implementation is built based on strict physical implications~\cite{phy1,phy2}, including a diffusion process and a reverse process. In the diffusion process, an image $\mathbf{X}_0$ is converted to a Gaussian distribution $\mathbf{X}_T$ by adding random Gaussian noise with $T$ iterations. The reverse process is to recover $\mathbf{X}_0$ from $\mathbf{X}_T$ by several denoising steps.

Abundant of recent diffusion methods focused on the task of text-to-image (T2I) generation. For instance, Glide~\cite{glid} proposed to combine the text feature into transformer blocks in the denoising process. Subsequently, DALL-E~\cite{t2i2}, Cogview~\cite{t2i3}, Make-a-scene~\cite{gafni2022make}, Stable Diffusion~\cite{ldm} and Imagen~\cite{t2i1} vastly improve the performance in T2I generation.
The widespread strategy is performing denoising in feature space and introducing the text condition into the denoising process by cross-attention model. 
Although they achieve promising synthesis quality, the text prompt can not provide the synthesis results with reliable structural guidance. PITI~\cite{piti} proposes to provide structural guidance by closing the distance between the feature of other types of conditions and the text condition. \cite{nvidia} proposes to utilize the similarity gradient between the target sketch and intermediate result to constrain the structure of the final results. Some methods~\cite{p2p,edit1,edit2} are also proposed to modulate the cross-attention maps in pre-trained T2I models to guide the generation process. One advantage of this type of approach is that they require no individual training. But they are still not practical in complex scenarios. As concurrent works, \cite{controlnet} learns task-specific ControlNet to enable conditional generation for the pre-trained T2I model. \cite{composer} proposed to retrain a diffusion model conditioned on a set of control factors. 

\subsection{Adapter}
The idea of adapter originated in the community of NLP. Adapter~\cite{adapter} found that it is not efficient to fine-tune a large pre-trained model for each downstream task and proposed transfer with an adapter, which is a compact and extensible model. \cite{pal} explored multi-task approaches that share a single BERT~\cite{bert} model with a small number of additional task-specific parameters. In the community of computer vision, \cite{plinvit} proposed to fine-tune the ViT~\cite{vit} for object detection with minimal adaptations. Recently, ViT-Adapter~\cite{vitadapter} utilized adapters to enable a plain ViT to perform different downstream tasks. However, the use of low-cost adapters on the pre-trained T2I model is still an open challenge.

\begin{figure*}[th]
\centering
\small 
\begin{minipage}[t]{\linewidth}
\centering
\includegraphics[width=1\columnwidth]{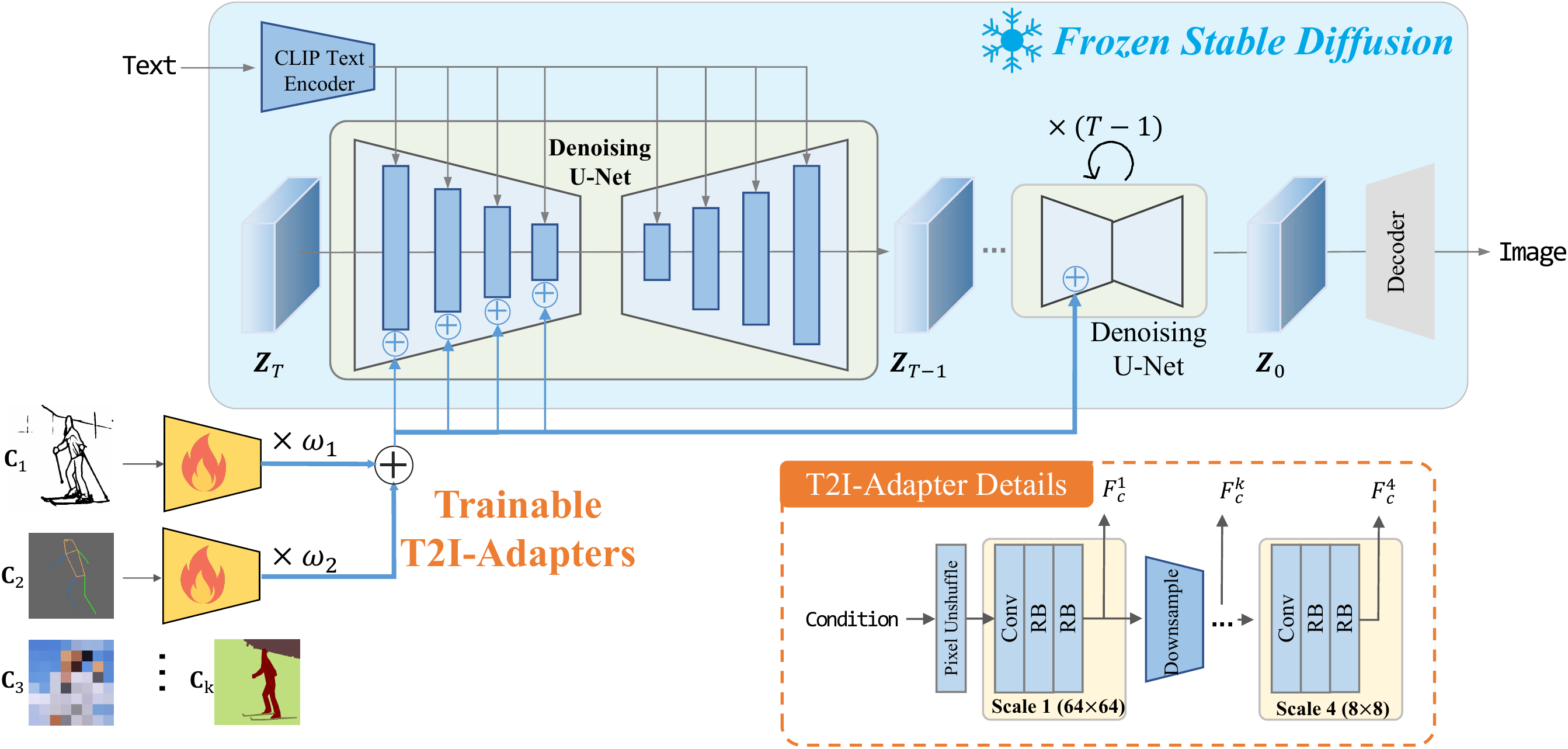}
\end{minipage}
\centering
\caption{The overall architecture is composed of two parts: 1) a pre-trained stable diffusion model with fixed parameters; 2) several T2I-Adapters trained to align internal knowledge in T2I models and external control signals. Different adapters can be composed by directly adding with adjustable weight $\omega$. The detailed architecture of T2I-Adapter is shown in the lower right corner.}
\vspace{-10pt}
\label{overview} 
\end{figure*}

\section{Method}
\subsection{Preliminary: Stable Diffusion}
In this paper, we implement our method based on the recent text-to-image diffusion model (\textit{i.e.}, Stable Diffusion (SD)~\cite{ldm}). 
SD is a two-stage diffusion model, which contains an autoencoder and an UNet denoiser. 
In the first stage, SD trained an autoencoder, which can convert images $\mathbf{X}_0$  into latent space and then reconstruct them. In the second stage, SD trained a modified UNet~\cite{unet} denoiser to directly perform denoising in the latent space. The optimization process can be defined as the following formulation:
\begin{equation}
    \mathcal{L} = \mathbb{E}_{\mathbf{Z}_{t},\mathbf{C},\mathbf{\epsilon},t}(||\mathbf{\epsilon} - \mathbf{\epsilon}_{\theta}(\mathbf{Z}_t,\mathbf{C})||_2^2),
\end{equation}
where $\mathbf{Z}_{t}=\sqrt{\overline{\alpha_{t}}}\mathbf{Z}_0+\sqrt{1-\overline{\alpha_{t}}}\mathbf{\epsilon},\ \mathbf{\epsilon} \in \mathcal{N}(0,\mathbf{I})$ represents the noised feature map at step $t$. $\mathbf{C}$ represents the conditional information. $\mathbf{\epsilon}_{\theta}$ refers to the function of UNet denoiser. During inference, the input latent map $\mathbf{Z}_T$ is generated from random Gaussian distribution.  Given $\mathbf{Z}_T$, $\mathbf{\epsilon}_{\theta}$ predicts a noise estimation at each step $t$, conditioned on $\mathbf{C}$. 
The noised feature map becomes progressively clearer by subtracting it. After T iterations, the final result $\hat{\mathbf{Z}}_0$, as the clean latent feature, is fed into the decoder of the autoencoder to perform image generation. In the conditional part, SD utilized the pre-trained CLIP~\cite{clip} text encoder to embed text inputs to a sequence of token $\mathbf{y}$. Then it utilizes the cross-attention model to combine $\mathbf{y}$ into the denoising process. It can be defined as the following formulation:
\begin{align}
\begin{split}
\left \{
\begin{array}{ll}
    \mathbf{Q} = \mathbf{W}_Q \phi(\mathbf{Z}_t);\ \mathbf{K} = \mathbf{W}_K \tau(\mathbf{y});\ \mathbf{V} = \mathbf{W}_V \tau(\mathbf{y})\\
    Attention(\mathbf{Q}, \mathbf{K}, \mathbf{V}) = softmax(\frac{\mathbf{Q}\mathbf{K}^T}{\sqrt{d}})\cdot \mathbf{V},
\end{array}
\right.
\end{split}
\end{align}
where $\phi(\cdot)$ and $\tau(\cdot)$ are two learnable embeddings. $\mathbf{W}_Q$, $\mathbf{W}_K$, and $\mathbf{W}_V$ are learnable projection matrices.

\begin{figure}[t]
\centering
\small 
\begin{minipage}[t]{\linewidth}
\centering
\includegraphics[width=1\columnwidth]{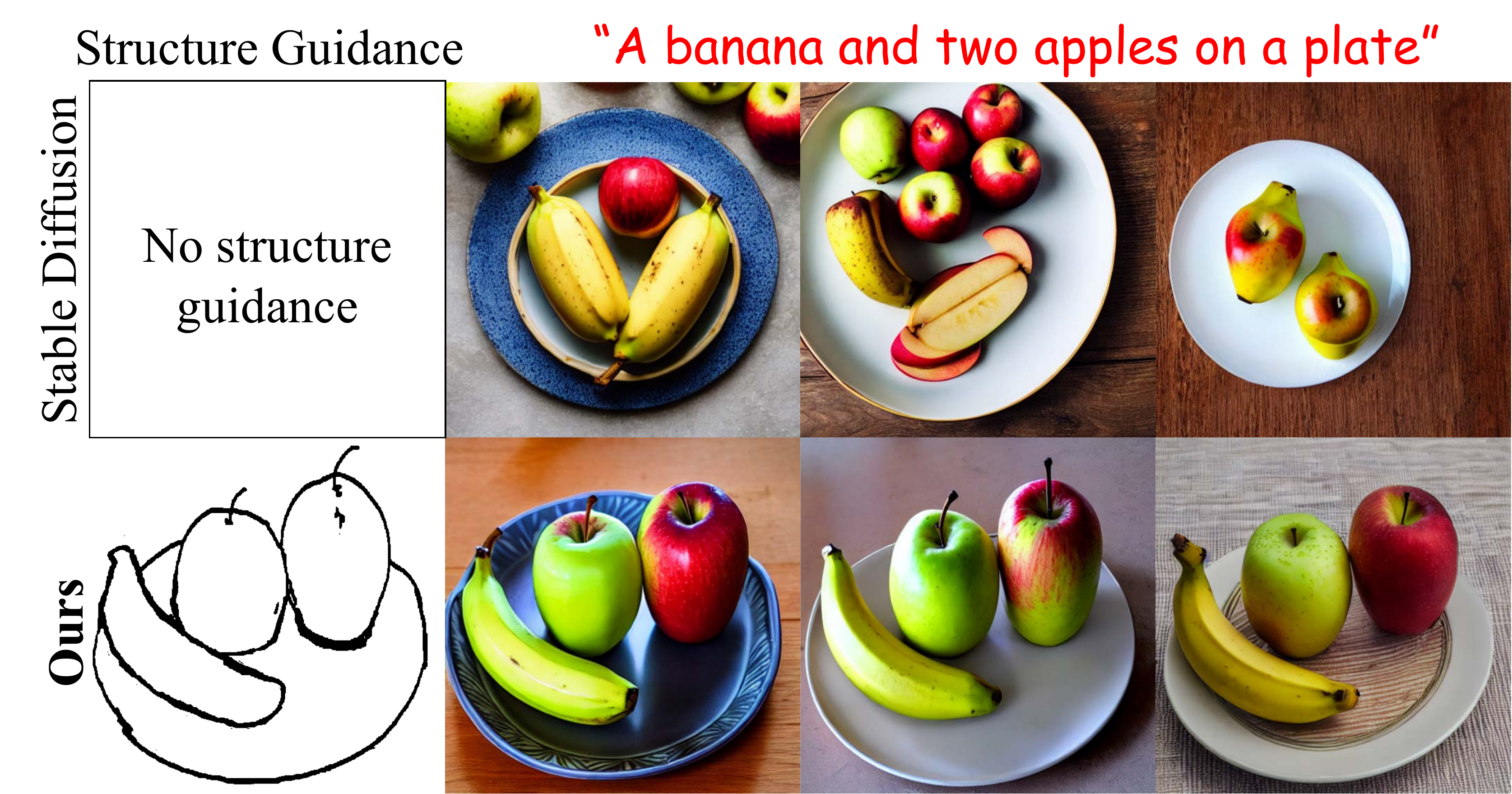}
\end{minipage}
\centering
\caption{In complex scenarios, SD fails to generate accurate results conforming to the text. In contrast, our T2I-Adapter can provide structure guidance to SD and generate plausible results.}
\vspace{-10pt}
\label{fail_sd} 
\end{figure}

\subsection{Overview of T2I-Adapter}
As shown in the first row of Fig.~\ref{fail_sd}, the text can hardly provide structural guidance to image synthesis, leading to random and unstable results in some complex scenarios. This is not due to poor generation ability, but because the text can not provide accurate generation guidance to fully align the internal knowledge of SD and external control. We believe that this alignment can be easily learned with low cost. An overview of our method is presented in Fig.~\ref{overview}, which is composed of a pre-trained SD model and several T2I adapters. 
The adapters are used to extract guidance features from different types of conditions. The pre-trained SD has fixed parameters to generate images based on the input text feature and extra guidance feature. 

\subsection{Adapter Design}
Our proposed T2I-adapter is simple and lightweight, as shown in the right corner of Fig.~\ref{overview}. It is composed of four feature extraction blocks and three downsample blocks to change the feature resolution. The original condition input has the resolution of $512\times 512$. Here, we utilize the pixel unshuffle~\cite{shuffle} operation to downsample it to $64\times 64$. In each scale, one convolution layer and two residual blocks (RB) are utilized to extract the condition feature $\mathbf{F}_{c}^k$. Finally, multi-scale condition features $\mathbf{F}_{c}=\{\mathbf{F}_c^1, \mathbf{F}_c^2, \mathbf{F}_c^3, \mathbf{F}_c^4\}$ are formed. Note that the dimension of $\mathbf{F}_{c}$ is the same as the intermediate feature $\mathbf{F}_{enc}=\{\mathbf{F}_{enc}^1, \mathbf{F}_{enc}^2, \mathbf{F}_{enc}^3, \mathbf{F}_{enc}^4\}$ in the encoder of UNet denoiser.
 $\mathbf{F}_c$ is then \textbf{added} with $\mathbf{F}_{enc}$ at each scale.
 In summary, the condition feature extraction and condition operation can be defined as the following formulation:
\begin{align}
    &\mathbf{F}_c = \mathcal{F}_{AD}(\mathbf{C})\\
    &\hat{\mathbf{F}}_{enc}^{i} = \mathbf{F}_{enc}^{i} + \mathbf{F}_{c}^{i},\ i\in \{1,2,3,4\}
\end{align}
where, $\mathbf{C}$ is the condition input. $\mathcal{F}_{AD}$ is the T2I adapter.

\noindent\textbf{Structure controlling.} Our proposed T2I-Adapter has a good generalization to support various structural control, including sketch, depth map, semantic segmentation map, and keypose. The condition maps of these modes are directly input into task-specific adapters to extract condition features $\mathbf{F}_{c}$.

\noindent \textbf{Spatial color palette.} In addition to structure, color is also a basic component of an image, which mainly involves two aspects: hue and spatial distribution. In this paper, we design a spatial color palette to roughly control the hue and color distribution of the generated images. To train the spatial palette, it is necessary to represent the hue and color distribution of an image. Here, we use high bicubic downsampling to remove the semantic and structural information of the image while preserving enough color information. Then we apply the nearest upsampling to restore the original size of the image. Finally, the hue and color distribution is represented by several spatial-arrangement color blocks. Empirically, we utilize $64\times$ downsampling and upsampling to complete this process. During training, we utilize the color map as $\mathbf{C}$ to generate $\mathbf{F}_c$ via $\mathcal{F}_{AD}$. 
 
\noindent\textbf{Multi-adapter controlling.} In addition to using a single adapter as a condition, our T2I adapters also support multiple conditions. Note that this strategy requires no extra training. Mathematically, this process can be defined as:
\begin{equation}
    \mathbf{F}_c = \sum_{k=1}^K \omega_k \mathcal{F}_{AD}^k(\mathbf{C}_k),
    \label{eq_compose}
\end{equation}
where $k\in [1, K]$ represents the $k$-th guidance. $\omega_k$ is the adjustable weight to control the composed strength of each adapter. This composable property leads to several useful applications. For instance, we can use the sketch map to provide structure guidance for the generated results while using the spatial color palette to color the generated results. More results are presented in Sec.~\ref{sec_comp}.

\begin{figure}[t]
\centering
\small 
\begin{minipage}[t]{\linewidth}
\centering
\includegraphics[width=1\columnwidth]{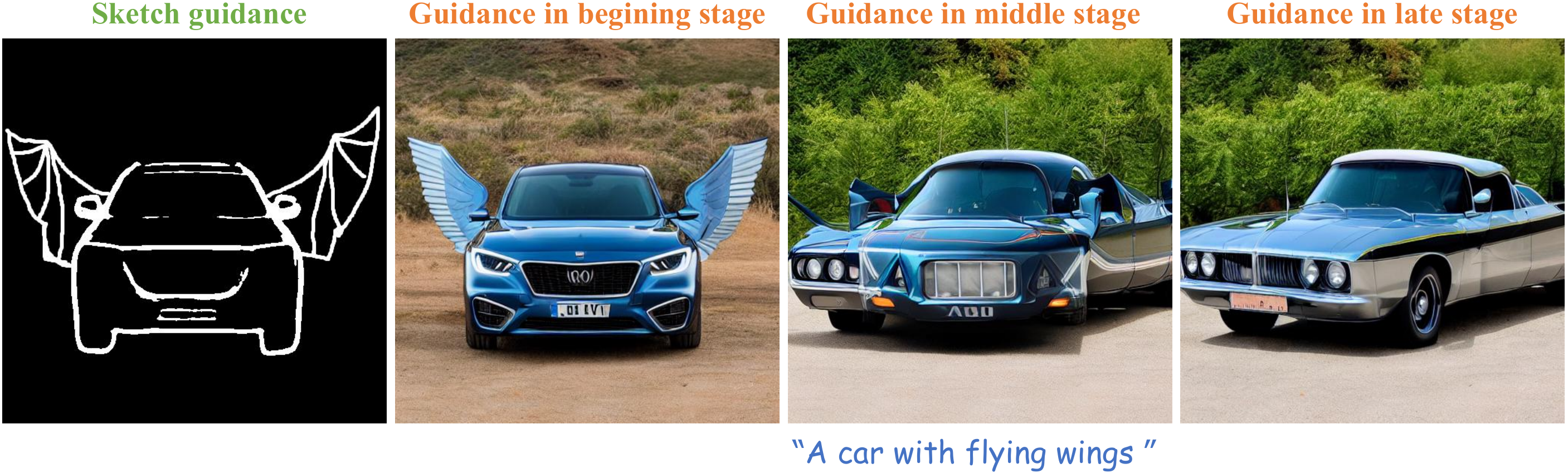}
\end{minipage}
\centering
\caption{
We evenly divide the DDIM inference sampling into 3 stages, \textit{i.e.}, beginning, middle and late stages. We observe the results of adding guidance at these three stages. Obviously, the later the iteration, the less the guiding effect.
}
\label{guidance_steps} 
\end{figure}

\begin{figure}[t]
\centering
\small 
\begin{minipage}[t]{\linewidth}
\centering
\includegraphics[width=1\columnwidth]{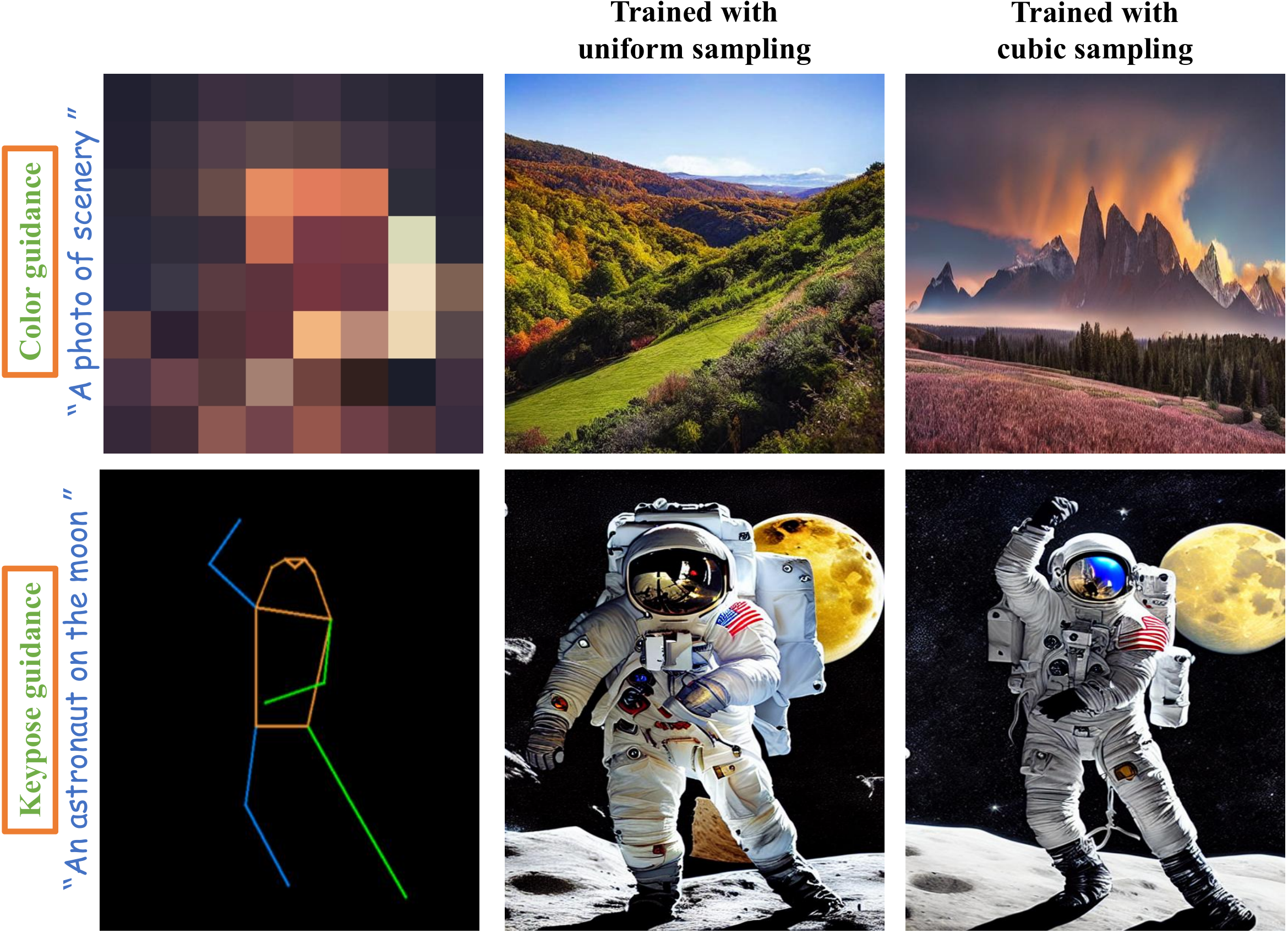}
\end{minipage}
\centering
\caption{The effect of cubic sampling during training. The uniform sampling of time steps has the problem of weak guidance, especially in color controlling. The cubic sampling strategy can rectify this weakness.}
\label{color_steps} 
\end{figure}


\begin{figure*}[t]
\centering
\small 
\begin{minipage}[t]{\linewidth}
\centering
\includegraphics[width=.95\columnwidth, height=6cm]{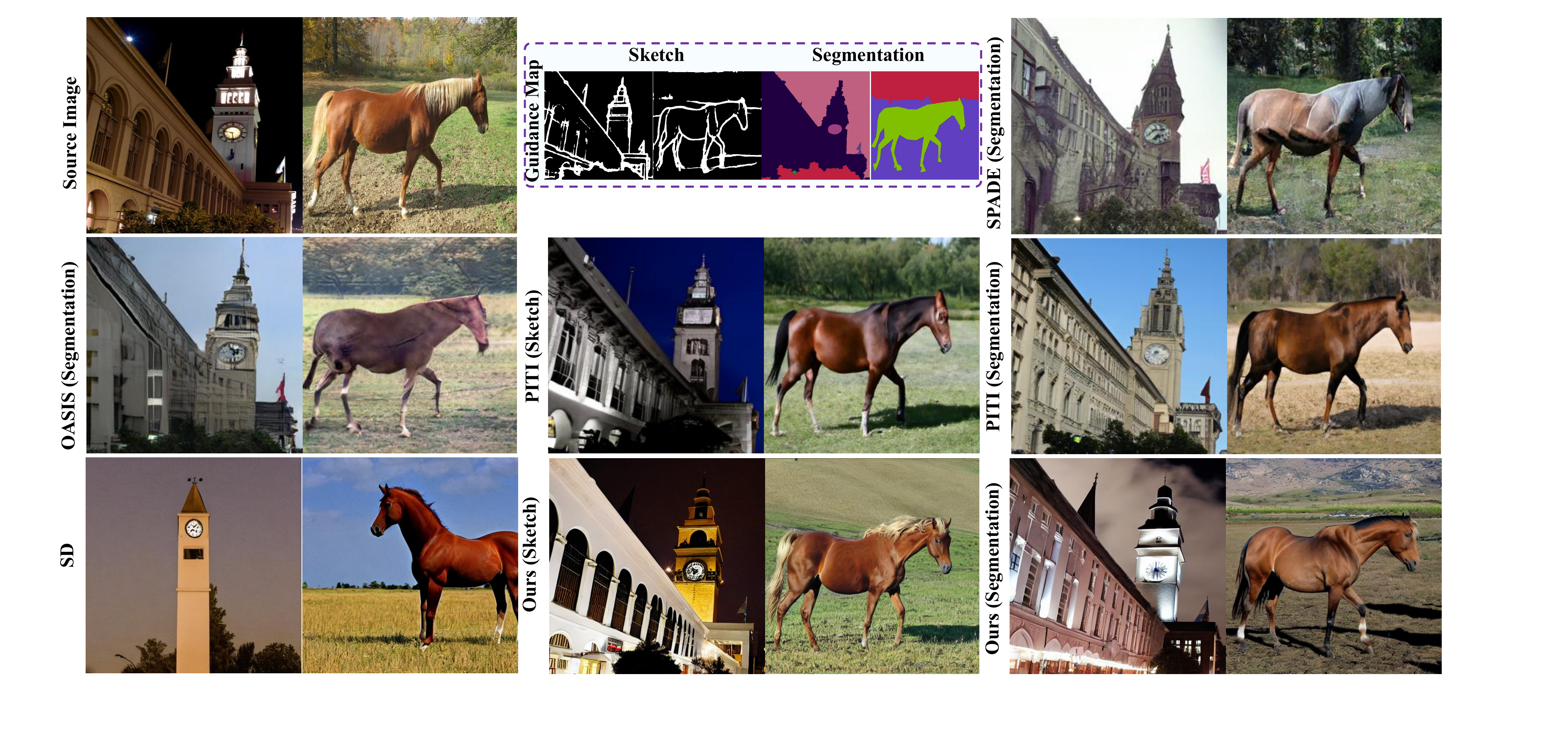}
\end{minipage}
\centering
\caption{The visualization comparsion between our method and other methods, \textit{i.e.}, SPADE~\cite{spade}, OASIS~\cite{oasis}, PITI~\cite{piti}, and SD~\cite{ldm}. Obviously, our method is superior to other methods in both alignment and generation quality.}
\label{compare_vis} 
\end{figure*}
\begin{table*}[t]
\caption{Quantitative comparison (FID~\cite{fid}/CLIP Score~\cite{clips} (ViT-L/14)) on COCO~\cite{coco} validation set. The best result is \textbf{highlighted}.
}
\footnotesize
\centering
\begin{tabular}{c c c c c c c c}
\toprule
 & \makecell[c]{SPADE~\cite{spade}\\ (segmentation)} & \makecell[c]{OASIS~\cite{oasis}\\ (segmentation)} & \makecell[c]{PITI~\cite{piti}\\ (segmentation)} & \makecell[c]{PITI~\cite{piti}\\ (sketch)} & \makecell[c]{SD~\cite{ldm}\\(text)} & \makecell[c]{Ours\\ (text+segmentation)} & \makecell[c]{Ours\\ (text+sketch)} \\ 
\hline
\hline
FID$\downarrow$ & 23.44 & 18.71 & 19.36 & 21.21 & 24.68 & \textbf{16.78} & 17.36\\ 
\hline
CLIP Score$\uparrow$ & 0.2314 & 0.2274 & 0.2287 & 0.2129 & 0.2648 & 0.2652 & \textbf{0.2666}\\ 
\toprule
\end{tabular}
\vspace{-10pt}
\label{cp_q1}
\end{table*}

\subsection{Model Optimization}
During optimization, we fix the parameters in SD and only optimize the T2I adapter. Each training sample is a triplet, including the original image $\mathbf{X}_0$, condition map $\mathbf{C}$, and text prompt $y$. The optimization process is similar to SD. Specifically, given an image $\mathbf{X}_0$, we first embed it to the latent space $\mathbf{Z}_0$ via the encoder of autoencoder. Then, we randomly sample a time step $t$ from $[0, T]$ and add corresponding noise to $\mathbf{Z}_0$, producing $\mathbf{Z}_t$. Mathematically, our T2I-Adapter is optimized via:
\begin{equation}
    \mathcal{L}_{AD} = \mathbb{E}_{\mathbf{Z}_{0},t, \textbf{F}_c, \epsilon \sim \mathcal{N}(0,1)}\left[ ||\epsilon-\epsilon_{\theta}(\mathbf{Z}_{t},t,\tau(\mathbf{y}),\textbf{F}_c)||_2^2\right]
\end{equation}

\noindent \textbf{Non-uniform time step sampling during training.} In the diffusion model, time embedding is an important condition in sampling. In our experiment, we also find that introducing time embedding into the adapter is helpful for enhancing the guidance ability. However, this design requires the adapter participating in each iteration, violating our motivation of simple and small. Therefore, we hope to rectify this weakness by suitable training strategies. There is an observation, shown in Fig.~\ref{guidance_steps}. Specifically, we evenly divide the DDIM inference sampling into 3 stages, \textit{i.e.}, beginning, middle and late stages. We then add guidance information to each of the three stages. We find that adding guidance in the middle and late stages had little effect on the result. It indicates that the main content of the generation results is determined in the early sampling stage. Therefore, the guidance information will be ignored during training if $t$ is sampled from the later section. To strengthen the training of adapter, non-uniform sampling is adopted to increase the probability of $t$ falling in the early sampling stage. Here, we utilize the cubic function (\textit{i.e.}, $t=(1-(\frac{t}{T})^3)\times T,\ t\in U(0,T)$) as the distribution of $t$. The comparison between uniform sampling and cubic sampling is presented in Fig.~\ref{color_steps}, including the color guidance and keypose guidance. We can find that the uniform sampling of $t$ has the problem of weak guidance, especially in the color controlling. The cubic sampling strategy can rectify this weakness.

\section{Experiment}
\subsection{Implementation Details}
We train T2I adapters for 10 epochs with a batch size of 8. We utilize Adam~\cite{adam} as the optimizer with the learning rate of $1\times 10^{-5}$. During training, we resize the input images and condition maps to $512\times 512$ and adapt the pre-trained SD model~\cite{ldm} with the version of 1.4. The training process is proformed on 4 NVIDIA Tesla 32G-V100 GPUs and can be completed within 3 days. 

Our experiment includes 5 types of conditions:
\begin{itemize}
    \item \textit{Sketch map.} For this structure condition, we utilize the training data of COCO17~\cite{coco}, contains $164K$ images, as the training dataset. The corresponding sketch maps are generated with the edge prediction model of~\cite{edge} and then are thresholded with $0.5$.
    \item \textit{Semantic segmentation map.} In this application, we utilize COCO-Stuff~\cite{coco_st} as the training data, which contains $164K$ images. Its semantic segmentation contains $80$ thing classes, $91$ stuff classes and $1$ \textit{`unlabeled'} class.
    \item \textit{Keypoints \& Color \& Depth maps.} For these applications, we select $600K$ images-text pairs from LAION-AESTHETICS dataset. For keypose, we use MMPose~\cite{mmpose} to extract the keypose map from each image. For depth, we utilize MiDaS~\cite{midas} to generate the depth maps of images.
\end{itemize}

\subsection{Comparison}
In this part, we select two commonly used generation guidances (\textit{i.e.}, sketch and segmentation) to compare our method with some state-of-the-art methods, \textit{e.g.}, some GAN-base~\cite{spade,oasis} and diffusion-based~\cite{piti} methods. The performance of original SD~\cite{ldm} is also evaluated. We utilize the COCO validation set, which contains $5,000$ images, to evaluate each method. For each image, different methods only randomly inference once as the final result. The visualize comparison is presented in Fig~\ref{compare_vis}. One can see that the result of our method is more vivid and more similar to the source image. The FID~\cite{fid} and CLIP Score (ViT-L/14)~\cite{clips} are applied as the quantitative evaluation to evaluate different method in Tab.~\ref{cp_q1}. The results show the promising performance of our method. We can also find that our T2I-Adapter not only brings regularity but also improves the performance of SD.

\begin{figure*}[t]
\centering
\small 
\begin{minipage}[t]{\linewidth}
\centering
\includegraphics[width=1\columnwidth]{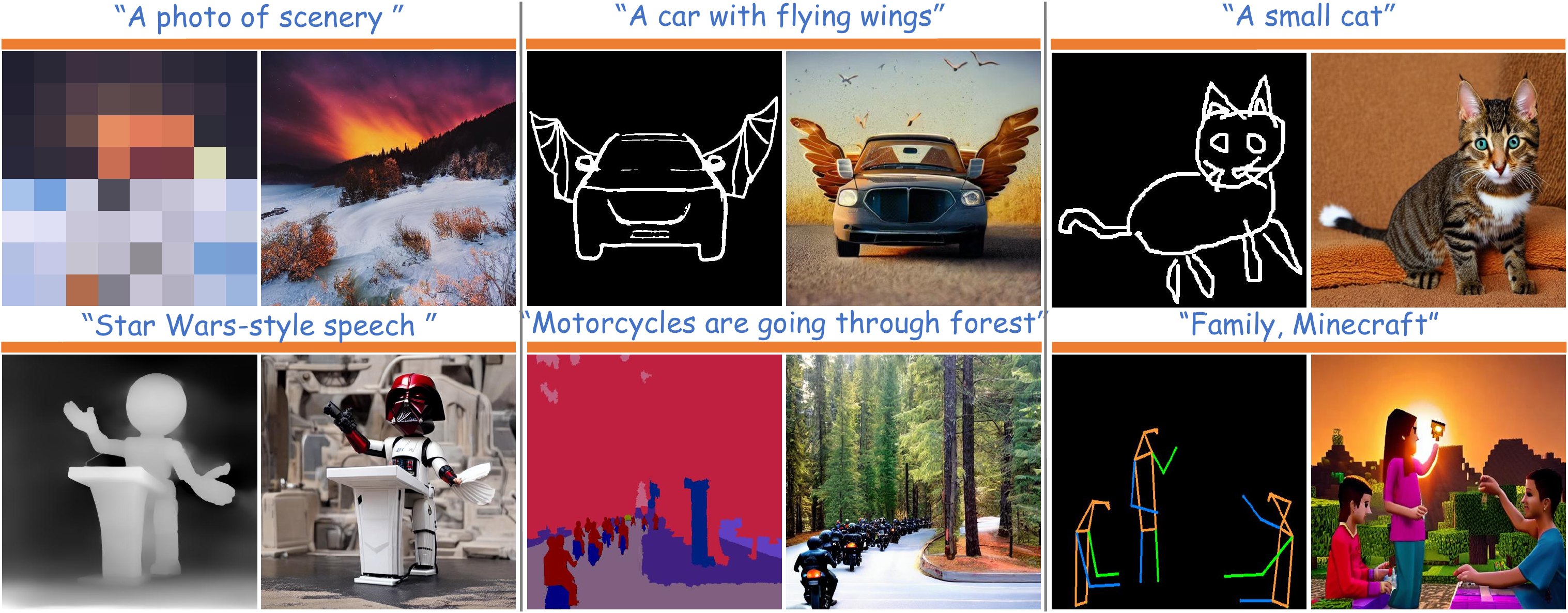}
\end{minipage}
\centering
\caption{Visualization of single-adapter controlling. With our proposed T2I-Adapter, the SD model can generate high-quality images conditioned on color map, sketch, depth map, semantic segmentation map, depth, and keypose.}
\vspace{-10pt}
\label{single_adapter} 
\end{figure*}

\subsection{Applications}
In this paper, we use several low-cost adapters to control the generation of a pre-trained T2I model, \textit{i.e.}, SD~\cite{ldm}. 
In this section, we will present several useful applications.

\subsubsection{Single-Adapter Controlling}
There are various control factors involved in our approach, including color, sketch, keypose, semantic segmentation, and depth. Fig.~\ref{single_adapter} presents the generation quality when using these guidance independently. One can see that they can play a corresponding controlling role on the generation, especially in some imaginative scenarios. 
At the same time, the adapter has good robustness, \textit{e.g.}, the small cat example which is a free-hand sketch. 

\begin{figure}[t]
\centering
\small 
\begin{minipage}[t]{\linewidth}
\centering
\includegraphics[width=1\columnwidth]{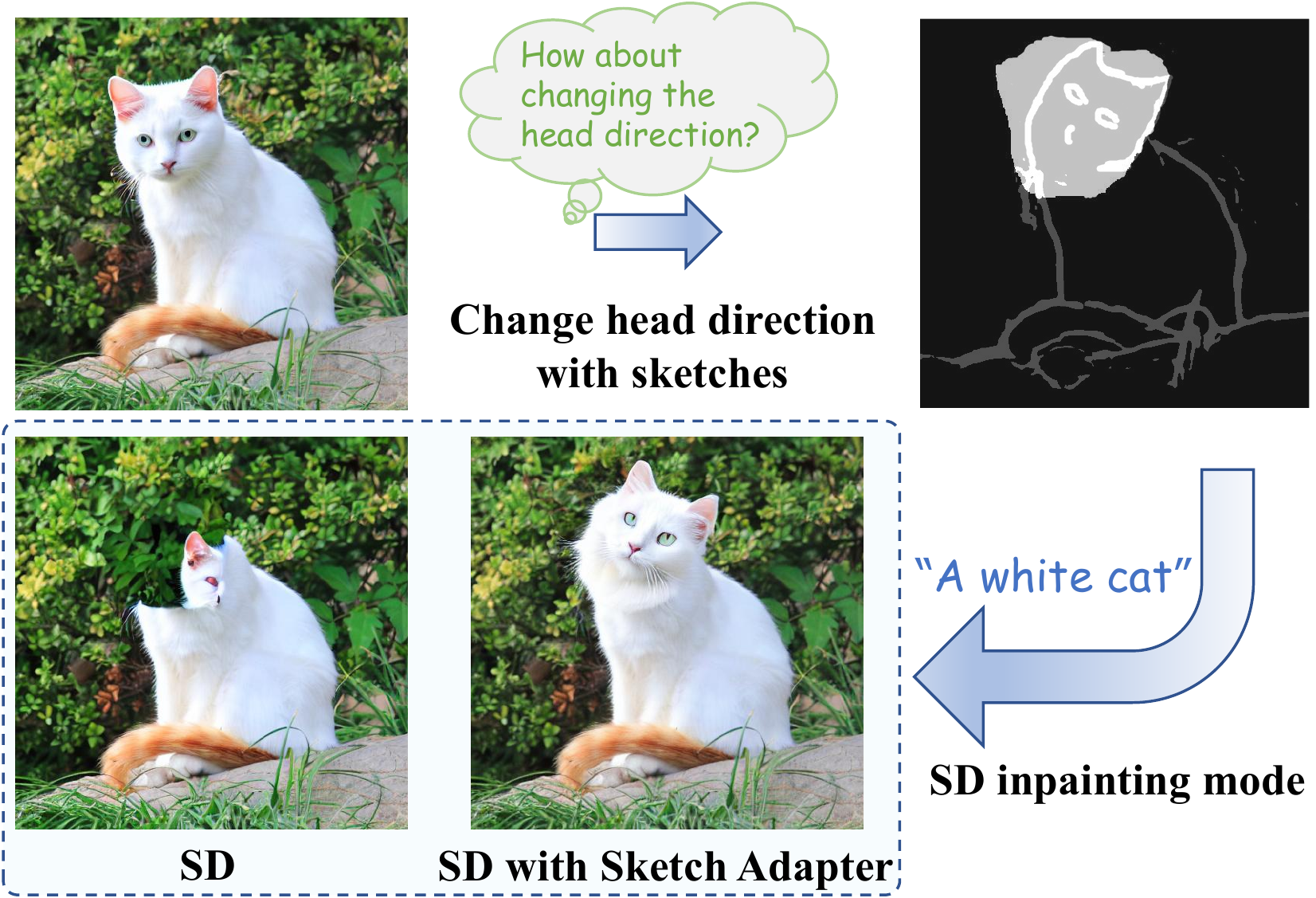}
\end{minipage}
\centering
\caption{Image editing ability of our sketch adapter. The inpainting result of SD~\cite{ldm} model is also presented as a comparison.}
\label{edit}
\vspace{-10pt}
\end{figure}

Based on the single-adapter controlling, we can also complete some image editing tasks, \textit{e.g.}, the result in Fig~\ref{edit}. Concretely, If the user is unsatisfied with a local region in an image, they can generate the desired content by erasing this area and inject the adapter guidance into the SD inpainting mode. In contrast, pure SD is difficult to achieve the same effect due to the ambiguous guidance of text.

\subsubsection{Composable Controlling}
\label{sec_comp}
\begin{figure}[h]
\centering
\small 
\begin{minipage}[t]{\linewidth}
\centering
\includegraphics[width=1\columnwidth]{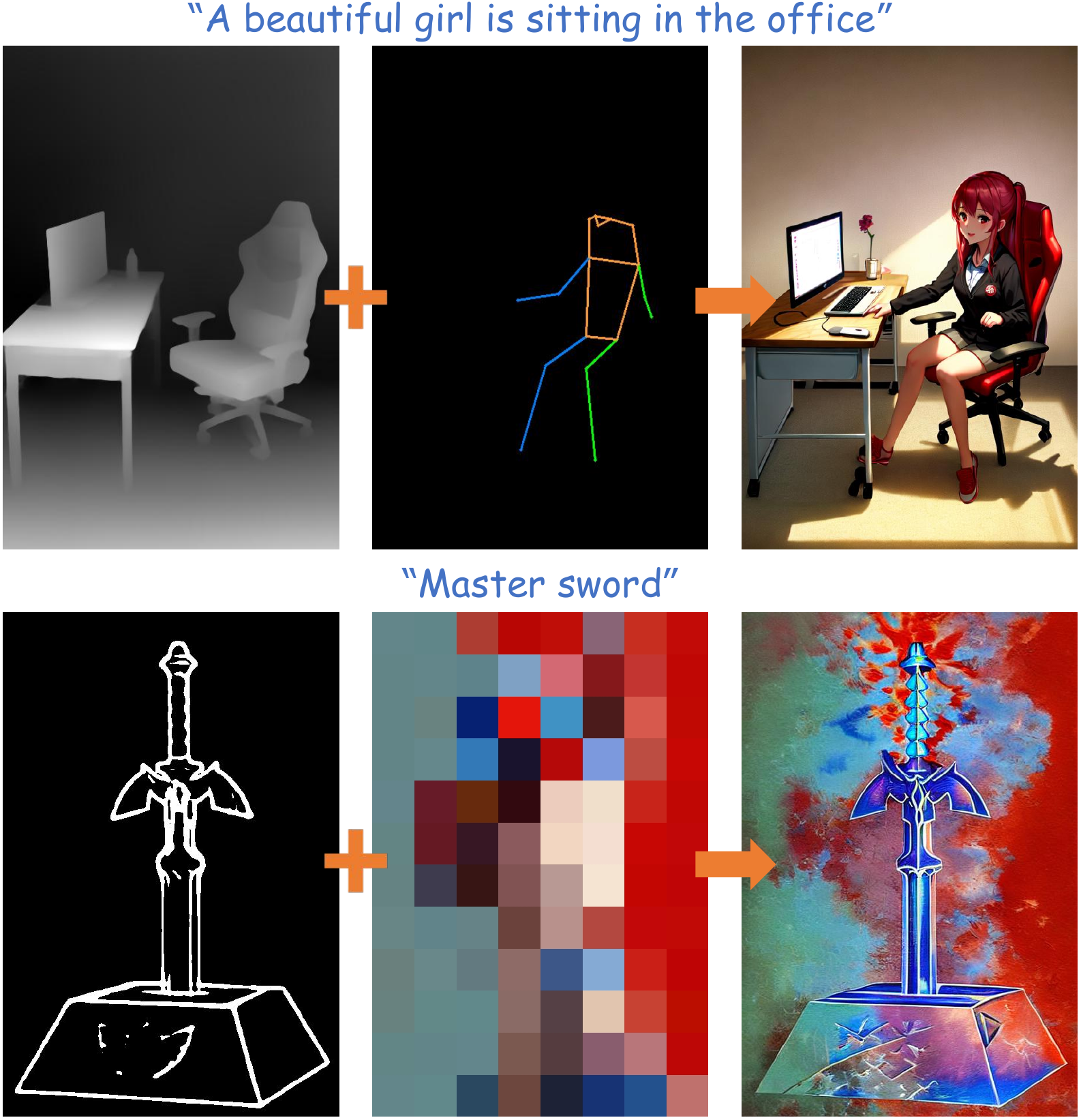}
\end{minipage}
\centering
\caption{Visualization of the composable controlling of our adapter, \textit{i.e.}, depth$+$keypose in the first row and sketch$+$color map in the second row.}
\label{compose_new}
\end{figure}

In addition to using these adapters individually, they can also be combined with each other without retraining, as shown in Eq.~\ref{eq_compose}. In Fig.~\ref{compose_new}, we present the results of depth adapter$+$keypose adapter and sketch adapter$+$color adapter. We can find that, there is a good composing and complementary ability between different adapters. 

\subsubsection{Generalizable Controlling}
\begin{figure}[t]
\centering
\small 
\begin{minipage}[t]{\linewidth}
\centering
\includegraphics[width=1\columnwidth]{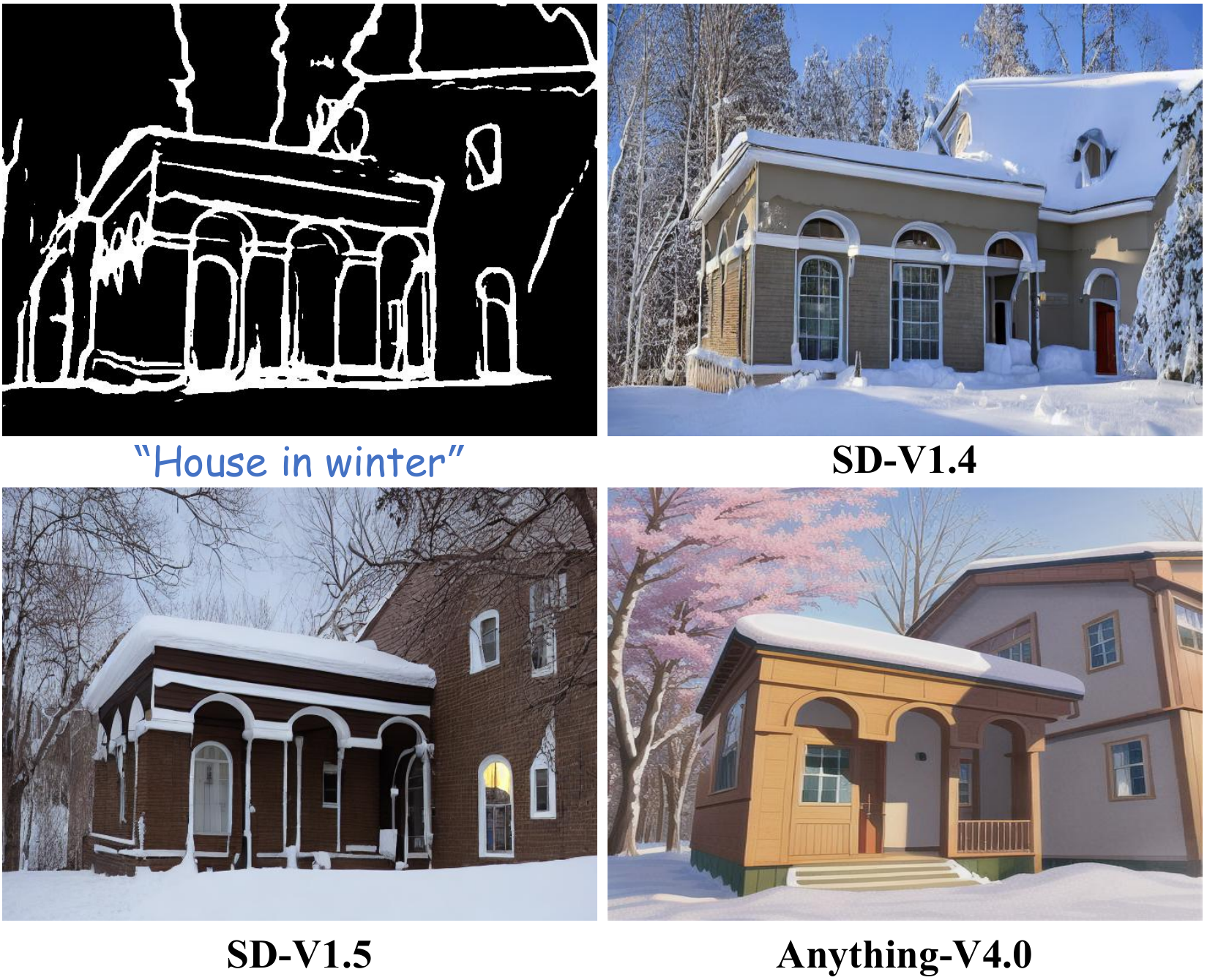}
\end{minipage}
\centering
\caption{Illustration of the generalizable ability of our T2I-Adapter. The sketch adapter is trained on SD-V1.4 and can perform well on SD-V1.5 and the custom model, \textit{e.g.}, Anything-V4.0.}
\label{general}
\end{figure}

The generalization ability of the adapter is an interesting and useful property. Concretely, once adapters are trained, they can be directly used on custom models as long as they are trained from the same T2I model. For instance, our adapters are trained on SD-V1.4, and they can perform the controlling on SD-V1.5 and other custom models~\footnote{https://huggingface.co/andite/anything-v4.0}, as shown in Fig.~\ref{general}. This generalization ability allows our T2I-Adapter to have a wider range of applications after a single training.

\begin{table}[t]
\caption{
Ablation study of how the guidance information is injected into the SD model. 
}
\centering
\begin{tabular}{c ||c c c ||c}
\toprule
Mode & \makecell[c]{Scale Num.} & Enc. & Dec. & FID \\ 
\hline
\hline
1 & 4 & \checkmark & \ding{55} & 17.36\\
2 & 4 & \ding{55} & \checkmark & 18.32\\
3 & 4 & \checkmark & \checkmark & 18.08\\
4 & 3 & \checkmark & \ding{55} & 17.86\\
5 & 2 & \checkmark & \ding{55} & 18.77\\
6 & 1 & \checkmark & \ding{55} & 22.66\\
\hline
\toprule
\end{tabular}
\label{tb_ablation}
\end{table}

\begin{figure}[t]
\centering
\small 
\begin{minipage}[t]{\linewidth}
\centering
\includegraphics[width=1\columnwidth]{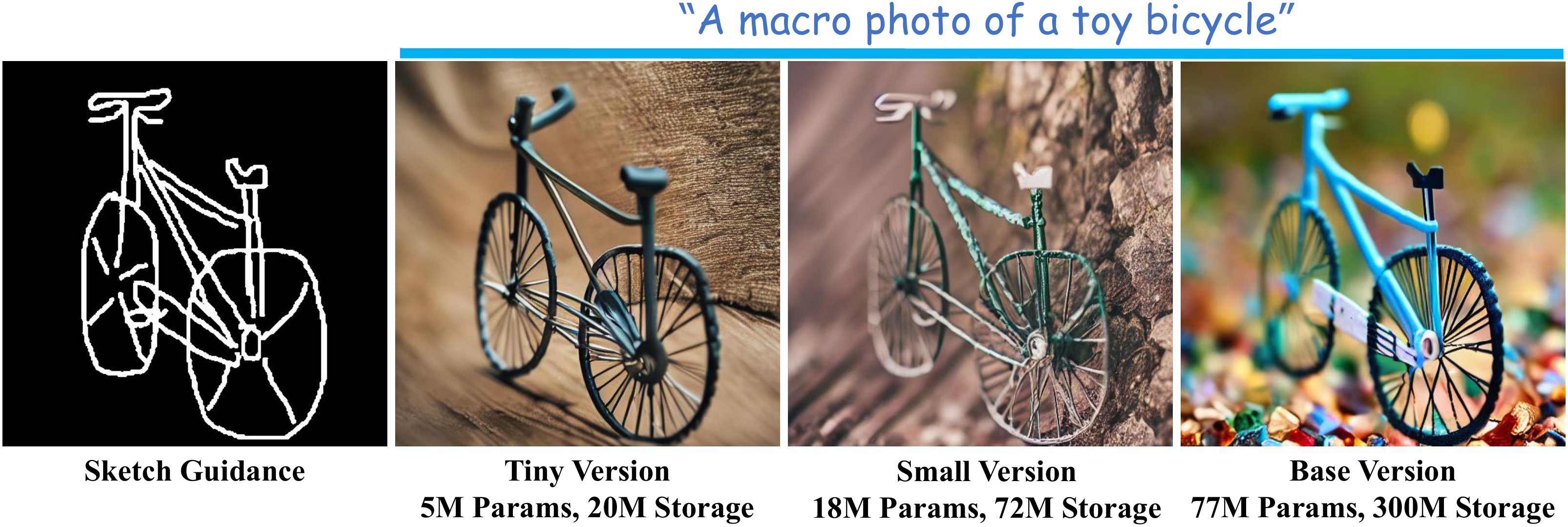}
\end{minipage}
\centering
\caption{The comparison of generation quality of the base, small, and tiny versions of our T2I-Adapter. All of them have attractive generation quality and control ability.}
\label{low_cost}
\end{figure}


\subsection{Ablation Study}
In this paper, we aim to utilize several low-cost and sample adapters to dig out more controllable ability from the SD model while not affecting their original network topology and generation ability. Therefore, in this part, we focus on studying the manner of injecting guidance information and the complexity of our T2I-Adapter. 

\subsubsection{Guidance Mode}
In this part, we study the manner of injecting guidance features into the SD model. The SD model has an encoder and a decoder, each with $4$ scales (\textit{i.e.}, $64\times 64$, $32\times 32$, $16\times 16$, $8\times 8$). Tab.~\ref{tb_ablation} presents the effect of injecting guidance information into these locations. Note that when the number of scales is less than 4, we preferentially discard the guidance feature with a small scale. One can see that it is more appropriate to inject guidance information into the encoder due to the longer information pathway of this strategy (containing the encoder and decoder). It can further refine the guidance feature obtained by our low-cost adapters. Multi-scale guidance features play a positive role in the generation results. We also found that injecting the guidance feature into both the encoder and decoder would cause much higher control strength, limiting the richness of texture. Finally, we chose to inject guidance features into all scales of the UNet encoder.

\subsubsection{Complexity Ablation}
Different from natural images, condition maps have higher sparsity. Therefore, we tend to use more lightweight models to extract these sparse features. In this part, we further compress the number of model parameters in the adapter by changing the channels of intermediate features, including $\times 4$ and $\times 8$ compression. Correspondingly, we get two smaller adapters, \textit{i.e.}, adapter-small (18M parameters) and adapter-tiny (5M parameters). Fig.~\ref{low_cost} presents the generation quality of these three versions. We can find that the tiny version still has attractive controlling capability in the sketch guidance. Considering that the color guidance is a more coarse-grained control compared with other structure guidance, our spatial color palette uses the small version of T2I-Adapter, and the base version is applied in other modes.

\section{Conclusion and Limitation}
In this paper, we aim to dig out the capabilities that T2I models have implicitly learned, \textit{e.g.}, the colorization and structuring capabilities, and then explicitly use them to control the generation more accurately. We present that a low-cost adapter model can achieve this purpose, as it is not learning new generation abilities but learning an alignment between the condition information and internal knowledge in pre-trained T2I models. In addition to the simplicity and lightweight structure, our T2I-Adapter 1) does not affect the original generation ability of the pre-trained T2I model; 2) has a wide range of applications in spatial color control and elaborate structure control. 3) More than one adapter can be easily composed to achieve multi-condition control. 4) Once trained, the T2I-Adapter can be directly used on custom models as long as they are fine-tuned from the same T2I model. Finally, extensive experiments demonstrate that the proposed T2I-Adapter  achieves excellent controlling and promising generation quality. One limitation of our method is that in the case of multi-adapter control, the combination of guidance features requires manual adjustment. In our future work, we will explore the adaptive fusion of multi-modal guidance information.


{\small
\bibliographystyle{ieee_fullname}
\bibliography{egbib}
}

\end{document}